%% file: submission.tex
\documentclass{article}
\usepackage{ijcai19}

\usepackage{times}
\usepackage{soul}
\usepackage{url}
\usepackage[hidelinks]{hyperref}
\usepackage[utf8]{inputenc}
\usepackage[small]{caption}
\usepackage{graphicx}
\usepackage{amsmath}
\usepackage{booktabs}
\usepackage{algorithm}
\usepackage{algorithmic}
\usepackage{natbib}

\input{math_commands.tex}

\title{Fairness in Reinforcement Learning}

\author{
    Paul Weng
    \affiliations
    Shanghai Jiao Tong University, Shanghai, China \\
    University of Michigan-Shanghai Jiao Tong University Joint Institute
    \emails
    paul.weng@sjtu.edu.cn
}

\begin{document}

\maketitle

\begin{abstract}
Decision support systems %(e.g., for ecological conservation) 
and autonomous systems %(e.g., adaptive controllers in smart cities) 
start to be deployed in real applications. 
Although their operations often impact many users or stakeholders, no fairness consideration is generally taken into account in their design, which could lead to completely unfair outcomes for some users or stakeholders. 
To tackle this issue, we advocate for the use of social welfare functions that encode fairness and present this general novel problem in the context of (deep) reinforcement learning, although it could possibly be extended to other machine learning tasks.
% propose an in-depth theoretical study of sequential
% decision-making under uncertainty with a welfare function (i.e., objective
% function that aggregates the utility of each user/stakeholder) that encodes
% fairness. 
% The difficulty of this new problem lies in the non-linearity of such an objective function, which changes the properties of optimal policies and prevents the direct application of dynamic programming or temporal difference methods. In
% this proposal, we investigate how properties (e.g., convexity, additive
% decomposability, etc) of fair welfare functions can be exploited to design
% efficient (possibly approximate) methods to find fair policies. More specifically,
% the main research content includes three aspects: (1) foundation for (adaptive)
% control, which provides theoretical results that will be used for designing
% algorithms for this new problem; (2) algorithms for fair control when the model is
% known; (3) algorithms for fair adaptive control when the model is unknown. Our
% final goal is to provide methods that can scale to large-sized problems, notably
% by using function approximation, such as deep neural networks. Furthermore, we
% will set up several experimental platform, notably traffic light control and data
% center control, to evaluate the proposed methods.
\end{abstract}

\section{Introduction}

Thanks to the progress in artificial intelligence and machine learning, but also notably to better sensors and increased computing power, decision support systems (DSS) and autonomous systems (AS) have started to become an integral part of our lives. 
A DSS %(e.g., multiple species management in ecological conservation \cite{Chades}) 
can help us make better, faster and more informed decisions in complex decision-making problems where generally multiple stakeholders are involved.
An AS %(e.g., smart traffic lights) 
can offer more efficient, more reactive and more adaptive control than human-operated systems or preprogrammed systems using fixed rules. 
However, as both DSS and AS are generally deployed among many users and may impact several stakeholders, fairness considerations become crucial for those systems to run successfully and to be accepted by all the different parties. 
%Indeed, a DSS is generally employed in complex decision-making problems with multiple stake-holders and a deployed AS often interacts with and affects many users. 
%Those users/stakeholders need to be equitably treated, otherwise the systems may be rejected. 
%Moreover, as advanced autonomous systems are typically very costly, they are sometimes funded and shared by several stakeholders (e.g., people, companies or countries). 
%Those stakeholders will only participate if they can receive a fair share from the usage of the systems. 
Thus, both systems need to be efficient in their solutions, but also fair to their users or stakeholders.

% From a social, economic and ecological point of view, those systems will help us face the problems related to a growing population, scarcer energy resources and pollution by making our lives safer, more comfortable, more productive and more frugal. 
% For instance, DSS used in ecological conservation
% for multiple species management helps preserve biodiversity \citep{XiaoDeeChadesPeyrardSabbadinStringerMcDonaldMadden18}.
% %Autonomous HVAC (heating, ventilation, and air conditioning) systems automatically manage the comfort of smart homes while reducing power consumption. 
% Intelligent traffic lights improve vehicle flow in smart cities by reducing the waiting times of cars and thus drivers’ frustration, but also by potentially decreasing fuel consumption and the number of accidents. 
% In such systems with many users/stakeholders (e.g., animal species or road users), fair optimization is important for their wide adoption and harmonious deployment. 
% %Moreover, fair solutions are key to enable the collaboration between different parties to build cutting-edge systems, such as satellites for Earth observation or telecom infrastructure, which are often shared by several entities (e.g., companies or countries). 
% In recent years, fairness has been recognized as a crucial consideration in resource allocation, notably in computer and telecommunication networks. 
% However, this notion and the research problems that it entails have not been much studied yet in machine learning, and especially deep reinforcement learning to the best of our knowledge.

Traditional (e.g., utilitarian) approaches  consist in optimizing a single cumulated cost/utility function (e.g., power consumption, QoS, QoE, financial and/or ecological cost) without any fairness consideration and are therefore insufficient, because in order to reach the overall optimum, the utility of some users/stakeholders could be unjustly sacrificed.
In order to take into account the welfare of each user/stakeholder, a multiobjective formulation, where each objective can be interpreted as the cost/utility of one user/stakeholder, is required. 
However, standard multiobjective methods generally focus on computing the set of Pareto-optimal solutions (solutions that cannot be improved on one objective, without worsening another). 
This is infeasible in practice because (1) this set may be extremely large, (2) in the case of AS, only one specific solution can be automatically applied and moreover, (3) Pareto-optimality itself does not encode any notion of fairness. 
An approach specifically designed for selecting a fair solution among the Pareto-optimal ones is therefore necessary.

As applications of artificial intelligence and machine learning start to pervade our everyday life, experts, policy makers and the general public start to realize that questions about fairness, ethics and safety are essential. 
Indeed, DSS and AS should not discriminate against us, should be designed to really help and not harm us. 
The problem presented in this paper fits in this new growing trend that proposes to enforce more human and social criteria to measure the quality of artificial systems. 
To achieve this goal, we describe an interdisciplinary approach that exploits results developed notably in economics (fairness models), applied mathematics (optimization and statistics) and computer science (machine learning).
For concreteness, we describe it in sequential decision-making problems. % in this paper.

% In order to design efficient algorithms to find fair solutions, the project starts with (1) the theoretical investigation of fair optimization in sequential decision-making under uncertainty, which will lay the foundation for the design of algorithms (2) in the case where the model of the environment is known and (3) in the case where it is unknown. 
% The general-purpose algorithms developed in this project, which will be designed so that they scale to large-sized problems, will pave the way for the practical design of autonomous systems evolving among many users or shared by several stakeholders. 
% Although the algorithms will be conceived to have a wide range of applications, they will be evaluated in this project on different specific application domains, notably the traffic light control problem and the data center control problem. 

\section{Background}

A sequential decision-making problem can be modeled as a Markov Decision Process (MDP).
In this section, we first recall this model and the reinforcement learning problem.
We then summarize multiobjective optimization approaches in sequential decision-making and underline their insufficiency for tackling fairness.
We finish this section with an overview of fairness modeling and fair optimization.
To simplify the presentation, we assume without loss of generality that the preferences of users/stakeholders are represented as utility (e.g., reward or payoff) to be maximized.

\paragraph{Markov Decision Process and Reinforcement Learning.}
 
% \begin{figure}[tb!]
% \centering
% \begin{tikzpicture}[node distance=1cm, auto]  
% \tikzset{
%     mynode/.style={rectangle,rounded corners,draw=black, top color=white, bottom color=yellow!50,very thick, inner sep=1em, minimum size=3em, text centered},
%     myarrow/.style={->, >=latex', shorten >=1pt, thick},
%     mylabel/.style={text width=7em, text centered} 
% }  
% \node[mynode] (Agent) {\iftoggle{chinese}{智能体}{Agent}};  
% \node[mynode, right=3cm of Agent] (Env) {\iftoggle{chinese}{环境}{Environment}};

% \draw[myarrow, bend left=50] (Agent.north) to node {\iftoggle{chinese}{当前状态的动作}{Action in Current State}} (Env.north);
% \draw[myarrow, bend left=50] (Env.south) to node {\iftoggle{chinese}{下个状态，即时奖励}{Next State, Immediate (Scalar) Reward}} (Agent.south);
% \end{tikzpicture} 
% \caption{\iftoggle{chinese}{智能体在环境中的规划／学习}{Planning/learning of an agent in an environment}}\label{fig:loop}
% \end{figure}

In an MDP \citep{Puterman94}, an agent repeatedly observes its state, chooses an action, obtains an immediate scalar numeric reward, and moves to a new state.
Solving an MDP (i.e., {\em planning}) amounts to finding a controller (called a {\em policy}) in order to maximize a standard decision criterion, e.g., the expected discounted reward or the expected average reward.
While in planning problems, the model of the environment (e.g., transition and reward functions) is assumed to be known, in reinforcement learning (RL) problems \citep{SuttonBarto98}, this assumption is relaxed:
an RL agent learns a best policy while interacting with the unknown environment by trial and error.

Thanks to their generality, those frameworks (MDP and RL) have been successfully applied in many diverse domains. 
For instance, MDPs and its extensions have been used for 
data center control \citep{WengQiuCostanzoYinSinopoli18} or
ecological conservation \citep{ChadesCarwardineMartinNicolSabbadinBuffet12}.
RL have been applied to 
robotics \citep{PetersVijayakumarSchaal03} or 
medicine \citep{PilarskiDawsonDegrisFahimiCareySutton11}.
The past few years, research in RL has become very active since the recent successes of the combination of deep learning and RL (called deep RL), notably in video games \citep{MnihKavukcuogluSilverRusuVenessBellemareGravesRiedmillerFidjelandOstrovskiPetersenBeattieSadikAntonoglouKingKumaranWierstraLeggHassabis15}.

% Although those frameworks are well-established, research is still active and on-going.
% Because of the curse of dimensionality, standard solving methods do not scale to large-sized domains that one can encounter in realistic problems.
% Therefore, one of the main research directions deals with the design of efficient methods that can scale to large-sized realistic problems, for example, 
% by exploiting structure (e.g., hierarchical decomposition in MDPs \citep{BaiWuChen15} or covering number in POMDPs \citep{ZhangHsuLeeLimBai15,ZhangHsuLee14,ZhangLittmanChen12}), 
% by using sampling techniques (e.g., trajectory sampling in MDPs \citep{ZhouLiuFuZhang15},  Thompson sampling in MDPs \citep{BaiWuZhangChen14}, or Monte Carlo Tree Search in POMDPs \citep{SilverVeness10}), 
% by relying on approximation (e.g., point-based solvers in POMDPs \citep{ShaniPineauKaplow13}, least-squares methods in RL \citep{XuHuLu07,XuHeHu02} or deep RL \citep{MnihKavukcuogluSilverRusuVenessBellemareGravesRiedmillerFidjelandOstrovskiPetersenBeattieSadikAntonoglouKingKumaranWierstraLeggHassabis15,MnihBadiaMirzaGravesLillicrapHarleySilverKavukcuoglu16,ZhangPanKochenderfer17}), or
% by using direct policy search (e.g., policy gradient \citep{SchulmanLevineAbbeelJordanMoritz15},  classification-based \citep{HuQianYu17} or
% evolutionary strategy \citep{SongYu14}).

\paragraph{Multiobjective Sequential Decision-making}

The standard models for sequential decision-making have been extended to the multiobjective (MO) setting \citep{RoijersVamplewWhitesonDazeley13,LiuXuHu15} (i.e., the immediate scalar numeric reward % in Figure~\ref{fig:loop} 
is replaced by a vector reward whose components represent objectives) where for instance, an objective can be interpreted in the multicriteria setting as a criterion (e.g., QoS, power consumption, monetary gain) to be optimized, or in the multi-user/stakeholder setting as the welfare of a user/stakeholder (e.g., average waiting times of car in different lanes for the traffic light control problem or QoS for different users for the data center control problem).
Most work in MO optimization (MOO) focuses on the multicriteria interpretation.
In this paper, we focus on the second interpretation, which naturally leads to fairness considerations (see the next part entitled \textit{Fair Optimization}).

The usual approach in MOO aims at finding the Pareto front, which is the set of all Pareto-optimal solutions   \citep{VamplewDazeleyBerryIssabekovDekker11}.
Unfortunately, computing the Pareto front is in general infeasible because the number of Pareto-optimal solutions can grow exponentially with the size of the problem \citep{PernyWengGoldsmithHanna13UAI}.
This observation may justify the computation of an approximation of those sets  \citep{LizotteBowlingMurphy10,PirottaParisiRestelli15}.
However, even with approximated sets, the approach is not suitable in autonomous systems where only one solution has to be automatically applied.

A solution to this issue relies on using a function that aggregates the objectives into a scalar value in order to select one solution among the set of Pareto-optimal solutions. 
However, one important point to realize is that the naive approach consisting in aggregating all the objectives with a weighted sum is insufficient.
Indeed, such a linear aggregation generally does not provide much control on the trade-offs between the objectives.
Moreover, non-supported (i.e., not on the convex hull) Pareto-optimal solutions cannot be obtained whatever the choice of the weights.

More interesting aggregating functions are non-linear and must be strictly increasing (in order to be monotonic with respect to Pareto dominance).
Such an approach is generally called compromise programming in the multicriteria setting, which generally consists in minimizing a distance to an ideal point \citep{Steuer86}.
Some work has been done for different functions in sequential decision making \citep{PernyWeng10ECAI,OgryczakPernyWeng13}.  %or weighted ordered weighted regret \citep{OgryczakPernyWeng11ADT}).
% Some of those results in compromise programming in MDPs will help for fair optimization, but we expect that the results developed in this project could also be adapted to compromise programming.
In the next paragraph, we present an aggregation function for modeling fairness, which we call fair welfare function.

\paragraph{Fair Optimization}

Fairness is a concept that has conventionally been studied in economics \citep{Moulin04} and political philosophy \citep{Rawls71}.
Recently, it has also become an important consideration in other applied fields, such as in applied mathematics \citep{OgryczakLussPioroNaceTomaszewski14} which focuses on solving fair optimization problems, in artificial intelligence \citep{de-JongTuylsVerbeeck08,HaoLeung16} when investigating multi-agent systems, or in computer engineering \citep{ShiPrasadOnurNiemegeers14} when designing computer networks.
As shown by recent surveys \citep{OgryczakLussPioroNaceTomaszewski14,Luss12}, fair optimization is an active and recent research area.
Although fairness is a key notion when dealing with multiple parties, 
it has only recently received attention in machine learning \citep{Busa-FeketeSzorenyiWengMannor17,SpeicherHeidariGrgicHlacaGummadiSinglaWellerZafar18,AgarwalBeygelzimerDudikLangfordWallach18,HeidariFerrariGummadiKrause18}.
To the best of our knowledge, the only work related to fairness in reinforcement learning investigate this issue in the multi-armed bandit setting \citep{Busa-FeketeSzorenyiWengMannor17}.

% fair optimization has not been  considered in machine learning\footnote{This approach is different to fair classification \citep{AgarwalBeygelzimerDudikLangfordWallach18} although their initial motivations share some similarity. We follow here the terminology used in economics and applied optimization.} to the best of our knowledge, except in a recent work in multi-armed bandits \citep{Busa-FeketeSzorenyiWengMannor17}.

Fairness can be defined in a theoretically-founded way \citep{Moulin04} and relies on two key principles.
The first one (P1) is called ``Equal treatment of equals'', which states that two users/stakeholders  (with identical characteristics with respect to the optimization problem, as assumed in this paper) should be treated the same way.
The second one (P2), called the {\em Pigou-Dalton principle}, is based on the notion of {\em Pigou-Dalton transfer}, which is a payoff transfer from a richer user/stakeholder to a poorer one without reversing their relative ranking.
The Pigou-Dalton principle states that such transfers lead to more equitable distributions.
Formally, for any $\bm v \in \mathbb R^n$ where $v_i < v_j$ and for any $\epsilon \in (0, v_j -v_i)$ we prefer $\bm v+\epsilon \mathbf 1_i - \epsilon \mathbf 1_j$ to $\bm v$ where $\mathbf 1_i$ (resp. $\mathbf 1_j$) is the canonical vector, null everywhere except in component $i$ (resp. $j$) where it is equal to $1$.
In words, this principle states that, all other things being equal, we prefer more ``balanced'' distributions (i.e., vectors) of payoffs.
Beside those two principles, as we are in an optimization context, an efficiency principle (P3) is also required, which states that given two payoff distributions, if one vector Pareto-dominates another, the former is preferred to the latter.

Those three principles imply that a fair welfare function that aggregates the payoffs of the users/stakeholders need to satisfy three properties.
They have to be symmetric (i.e., independent to the order of its arguments for P1), strictly Schur-concave (i.e., monotonic with respect to Pigou-Dalton transfers for P2) and strictly increasing (i.e., monotonic with respect to Pareto dominance for P3).
The elementary approach based on maximin (or Egalitarian approach), where one aims at maximizing the worse-off user/stakeholder, does not satisfy the last two properties.
A better approach \citep{Rawls71} is based on the lexicographic maximin, which consists in comparing first the worse-off user/stakeholder when comparing two vectors, then in case of a tie, comparing the second worse-off and so on.
However, due to the non-compensatory nature of the min operator, vector $(1, 1, \ldots, 1)$ would be preferred to $(0, 100, \ldots, 100)$, which may be debatable.

% Two families of fair welfare functions that satisfied all three properties have been particularly studied:
% \begin{itemize}
% \item Sum of increasing strictly concave function \citep{KOWejor04} $F_{u} : \mathbb R^n \to \mathbb R$:
% \begin{align} \label{eq:fu}
% F_u(\bm v) = \sum_i u(v_i)
% \end{align}
% where $u : \mathbb R \to \mathbb R$ is an increasing strictly concave function, $\bm v = (v_1, v_2, \ldots, v_n)$ is a payoff distribution vector, and $n$ is the number of users/stakeholders.

% Function $F_u$ defines a very general family of welfare functions.
% It can notably represent proportional fairness \citep{PioroMalicskoFodor02} when $u(x) = \log(x)$ and more generally $\alpha$-fairness \citep{MoWalrand00} when $u_\alpha(x) = \frac{x^{1-\alpha}}{1-\alpha}$ if $\alpha \neq 1$ and $u_\alpha(x) = \log(x)$ otherwise, with a parameter $\alpha>0$ controlling the aversion to inequality.
% When $\alpha\to\infty$, $F_{u_\alpha}$ tends to lexicographic maxmin.
% Even more broadly, this family includes welfare functions derived from the generalized entropy index \citep{Shorrocks80}.

%\item 
Many fair welfare function have been proposed.
In practice, the choice of a suitable function depends on the application domain.
For illustration, we present the
fair welfare function based on the Generalized Gini Index (GGI) \citep{Weymark81} $G_{\bm w} : \mathbb R^n \to \mathbb R$:
\begin{align}\label{eq:gw}
G_{\bm w}(\bm v) = \sum_{i} w_i v^\uparrow_i
\end{align}
where $\bm w \in [0, 1]^n$ is a weight vector such as $w_1 > w_2 > \ldots > w_n$, and $(v^\uparrow_1, v^\uparrow_2, \ldots, v^\uparrow_n)$ is the payoff vector $\bm v$ reordered in an increasing fashion.

Functions $G_{\bm w}$ contains the welfare function induced by the classic Gini index or the Bonferroni index \citep{Tarsitano90}.
It tends to the Egalitarian approach when $w_2 \to 0$, $\ldots$, $w_n \to 0$ and to the lexicographic maxmin when differences between weights tends to infinity.
% \end{itemize}

%Both families of welfare function can be extended to give different importance to different users/stakeholders.
%The methods developed in this project could readily be extended to this case, therefore we do not specifically discuss it in more details.
%Both $F_u$ and $G_{\bm w}$ have been used in numerous application domains.
%The first family $F_u$ has been considered for instance, in network protocol design \citep{MoWalrand00}, more generally in network resource allocation \citep{LanKaoChiangSabharwal10}, and in wireless networks \citep{ShiPrasadOnurNiemegeers14}.
%It has been used for optimization in traffic allocation problems \citep{KellyMaullooTan97} and link capacity dimensioning \citep{PioroMalicskoFodor02}.
%The second family $G_{\bm w}$ 
GGI has been exploited in different MO (continuous and combinatorial) optimization problems.
To cite a few, it was used %in network dimensioning problems \citep{OgryczakSliwinskiWierzbicki03},
in capital budgeting \citep{KOWejor04}, allocation problems \citep{NguyenWeng17}, or
flow optimization in wireless mesh networks \citep{HurkalaSliwinski12}.

\section{Problem Formulation}
%\subsection{Problem Formulation}

At a high-level, a fair sequential decision-making problem can be understood as solving a non-linear convex optimization problem\footnote{The objective function is convex when minimizing costs and concave when maximizing utilities. As customary in the optimization literature, we may refer to both problems as convex optimization problems.}, where the welfare function, which encodes both efficiency and fairness, aggregates the utility of each user/stakeholder:
\begin{align}
\max_\pi ~& J(\pi) = H( \sum_s \mu(s) \bm V^\pi(s) )\label{eq:go}
\end{align}
where $\pi$ is a policy, 
$H$ is a fair welfare function (e.g., GGI), % (i.e., $F_u$ or $G_{\bm w}$), 
$\mu$ is a probability distribution over initial states, and $\bm V^\pi$ is the multiobjective value function of $\pi$ (e.g., expected discounted or average reward).

%\subsection{Difficulties}
The difficulty of this new problem lies in the non-linearity of the objective function, which changes the properties of optimal policies and prevents the direct application of dynamic programming or temporal difference methods. 
However, the properties (e.g., concavity, Schur-concavity, decomposability, etc) of fair welfare functions and those (e.g., temporal structure) of sequential decision-making problems can be exploited to design efficient methods to find fair policies. 
%More specifically,
% the main research content includes three aspects: (1) foundation for (adaptive)
% control, which provides theoretical results that will be used for designing
% algorithms for this new problem; (2) algorithms for fair control when the model is
% known; (3) algorithms for fair adaptive control when the model is unknown. Our
% final goal is to provide methods that can scale to large-sized problems, notably
% by using function approximation, such as deep neural networks. Furthermore, we
% will set up several experimental platform, notably traffic light control and data
% center control, to evaluate the proposed methods.

\section{Preliminary Experimental Results}

\begin{figure}[tb]
\centering 
\includegraphics[width=.4\textwidth]{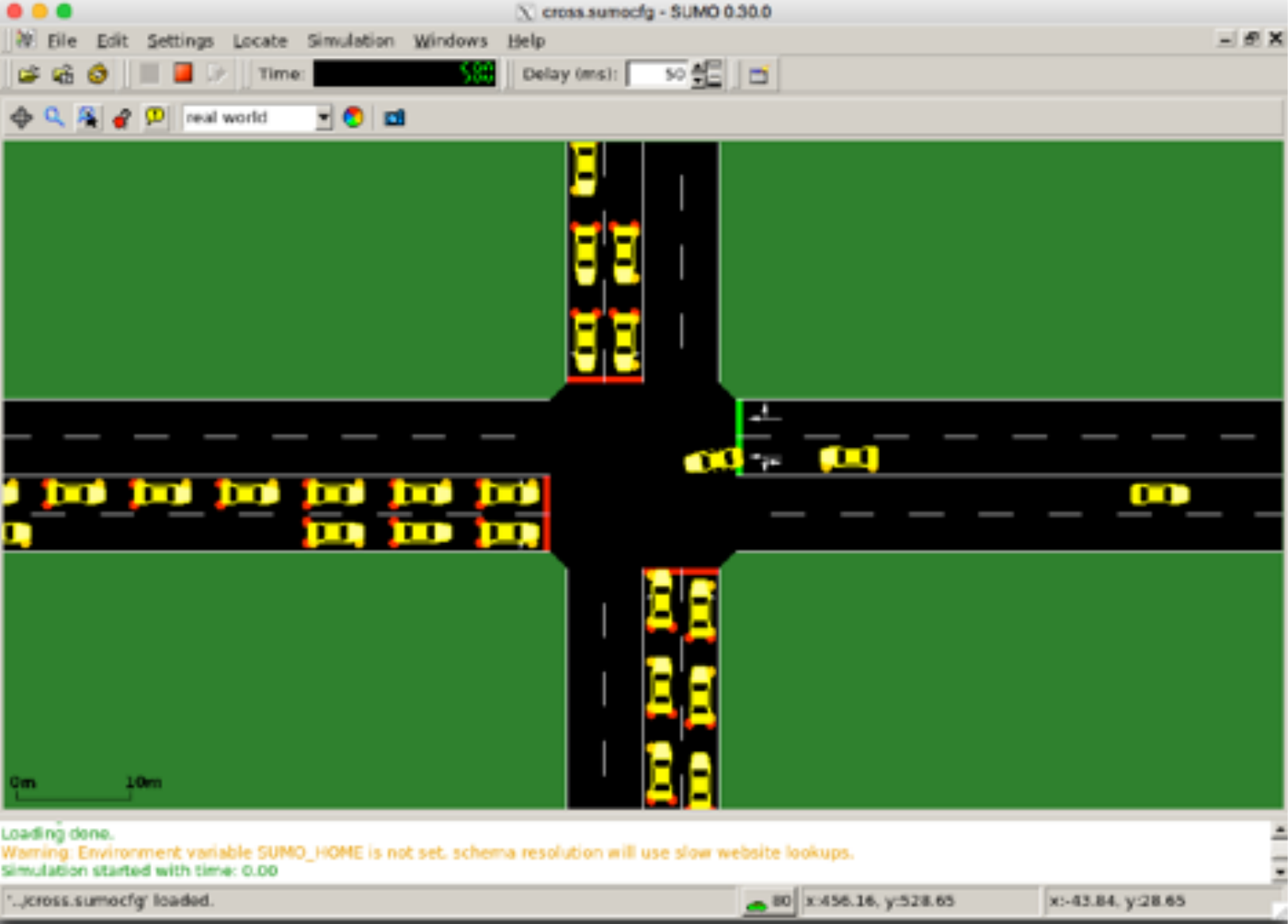}
\hfill
\includegraphics[width=.4\textwidth,height=4.8cm]{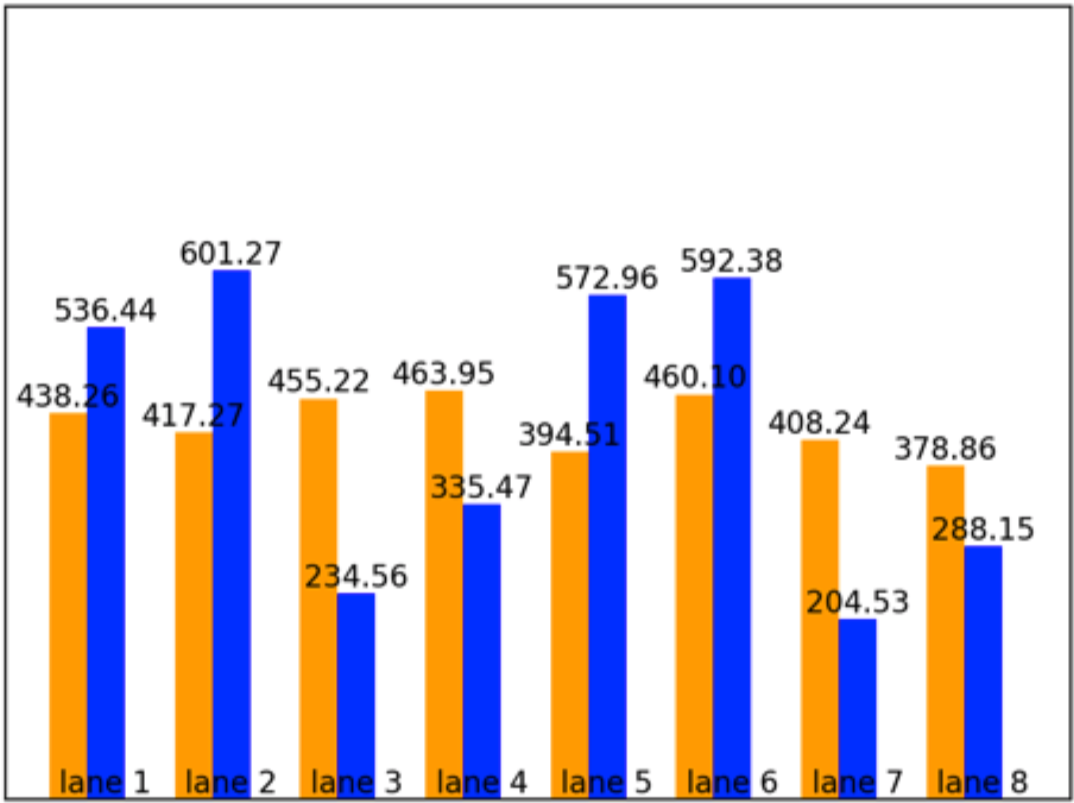}
\caption{Left: Screenshot of the SUMO simulator; Right: Average waiting times for standard DQN (blue) vs GGI-DQN (orange).}
\label{fig:intersection wt}
\end{figure}

To demonstrate the potential usefulness of our proposition, we conducted some initial experiments in a traffic light control problem, because such environments are relatively easy to simulate.
We use SUMO\footnote{\url{http://sumo.dlr.de/index.html}} (see on the top of Fig.~\ref{fig:intersection wt} for an illustration) to simulate one intersection with a total of 8 lanes under varying traffic conditions.
Standard approaches to solve this problem usually minimize the expected waiting times over all lanes.
In our formulation, we learn a traffic controller that attempts to minimize the expected waiting times of each lane, while ensuring some notion of fairness over each lane is enforced.
In our experiments, we used the generalized Gini index and adapted the DQN algorithm \citep{MnihKavukcuogluSilverRusuVenessBellemareGravesRiedmillerFidjelandOstrovskiPetersenBeattieSadikAntonoglouKingKumaranWierstraLeggHassabis15} to approximately optimize it.
Although we illustrate the approach on the traffic light domain, the method could be applied to diverse other sequential-decision-making problems.

Fig.~\ref{fig:intersection wt} (bottom) shows some initial results (averaged over 20 runs) where we compare our proposed approach (GGI-DQN in orange) with the standard approach (DQN in blue) that minimizes the expected waiting times over all lanes.
As expected DQN obtains a lower average waiting times over all lanes (as it optimizes this criterion) than GGI-DQN: 420.72 vs 427.05 (in timesteps in the simulator).
However, the average waiting times \textit{in} each lane for the standard approach have an unequal distribution, while
our approach provides a much fairer distribution of waiting times.

\section{Conclusion}

In this paper, we argued for the use of fair welfare functions in machine learning tasks and demonstrated it more specifically in reinforcement learning.
%We presented fair optimization in the context of sequential decision-making problems.
We believe that the topic of fair optimization is novel in machine learning and is of great significance, as it naturally provides solutions that take into account the welfare of all the involved parties.
As future work, we plan to develop more efficient algorithms in the deep RL setting for optimizing different fair welfare functions, and possibly extend the approach to other machine learning tasks.
%Given the importance of autonomous systems, the advantage that they can bring and the inadequacy of the existing methods to take into account the interests of multiple parties, the topic of fair optimization for adaptive control is novel and of great significance.

% Moreover, the results obtained through the basic research of this project will open new research directions not considered in this proposal, such as fair distributed control, incremental preference elicitation of the welfare function, or strategy-proofness with respect to the user/stakeholder in control problems.

\subsubsection*{Acknowledgments}

We would like to thank the anonymous reviewers for their helpful feedback.
This work is supported by the Shanghai NSF (No. 19ZR1426700) and in part by the program of National Natural Science Foundation of China (No. 61872238).

\bibliography{biblio190214}
\bibliographystyle{named}

\end{document}

%% file: math_commands.tex
\usepackage{amsmath,amsfonts,bm}

\def\eqref#1{equation~\ref{#1}}
\def\1{\bm{1}}

\DeclareMathAlphabet{\mathsfit}{\encodingdefault}{\sfdefault}{m}{sl}
\SetMathAlphabet{\mathsfit}{bold}{\encodingdefault}{\sfdefault}{bx}{n}